\documentclass[%
 aip,
 floatfix,
 amsmath,amssymb,
 reprint,%
]{revtex4-1}
\usepackage{graphicx}
\usepackage{setspace}

\usepackage{hyperref}
\hypersetup{
    colorlinks=true,
    linkcolor=blue,
    filecolor=magenta,      
    urlcolor=blue,
    pdftitle={Overleaf Example},
    citecolor=green,
    pdfpagemode=FullScreen,
    }

\makeatletter
\def\@email#1#2{%
 \endgroup
 \patchcmd{\titleblock@produce}
  {\frontmatter@RRAPformat}
  {\frontmatter@RRAPformat{\produce@RRAP{*#1\href{mailto:#2}{#2}}}\frontmatter@RRAPformat}
  {}{}
}%
\makeatother

\begin{document}

\preprint{AIP/123-QED}

\title[MeshDQN]{MeshDQN: A Deep Reinforcement Learning Framework for Improving Meshes in Computational Fluid Dynamics}

\author{Cooper Lorsung}
\author{Amir Barati Farimani}%
\altaffiliation[Also at ]{Department of Chemical Engineering, Carnegie Mellon University, Pittsburgh, PA 15213, USA}
\affiliation{ 
Department of Mechanical Engineering, Carnegie Mellon University, Pittsburgh, PA 15213, USA
}%
\date{\today}

\begin{abstract}
    Meshing is a critical, but user-intensive process necessary for stable and accurate simulations in computational fluid dynamics (CFD).
    Mesh generation is often a bottleneck in CFD pipelines.
    Adaptive meshing techniques allow the mesh to be updated automatically to produce an accurate solution for the problem at hand.
    Existing classical techniques for adaptive meshing require either additional functionality out of solvers, many training simulations, or both. 
    Current machine learning techniques often require substantial computational cost for training data generation, and are restricted in scope to the training data flow regime.
    MeshDQN is developed as a general purpose deep reinforcement learning framework to iteratively coarsen meshes while preserving target property calculation.
    A graph neural network based deep Q network is used to select mesh vertices for removal and solution interpolation is used to bypass expensive simulations at each step in the improvement process.
    MeshDQN requires a single simulation prior to mesh coarsening, while making no assumptions about flow regime, mesh type, or solver, only requiring the ability to modify meshes directly in a CFD pipeline.
    MeshDQN successfully improves meshes for two 2D airfoils.
\end{abstract}


\maketitle

\section{Introduction}

Mesh generation in computational fluid dynamics (CFD) is a critical step for stable and accurate simulations.
However, one of seven key findings from a year-long NASA study published in 2014 is that mesh generation and adaptation remains a significant bottleneck in the CFD workflow\cite{NASA}.
One of the reasons for this is meshing tends to be done with intuition and experience, rather than a strictly mathematical approach, to get convergence and stable simulations\cite{BAKER200529}.
In order to achieve the goals of the NASA study, autonomous mesh generation is a necessary next step.
A more recent review on the progress of modeling and mesh generation since the study has indicated a growing interest in adaptive mesh techniques to aid in mesh generation \cite{chawner_progress_2019}.

Substantial literature exists on using both classical as well as machine learning techniques for adaptive meshing\cite{park_unstructured_2016, ALAUZET201613, ml_optimal_mesh}.
Much of the focus on classical approaches is on pointwise error estimation and subsequent refinement\cite{ALAUZET201613}.
While these methods can be used to improve meshes, they often require additional backward solves to compute adjoint states with a full simulation at each step to recalculate the velocity and pressure fields.
This adds computational time, as well as additional necessary features into a CFD solver, often rendering the method unfeasible with CFD codes and pipelines.
More recently, many machine learning models and methods have been developed for adaptive meshing.
However, these methods tend to rely on substantial up-front computational time in training data generation, where many simulations are run\cite{supermeshing}, and adjoint-based adaptive meshing are used\cite{ml_optimal_mesh}.

Iterative mesh refinement can instead be viewed as a sequential decision making process, where each region or vertex to be modified constitutes another decision.
This is well suited for deep reinforcement learning (DRL).
Recent surveys have shown DRL applied to CFD has been primarily focused on control and optimization\cite{drl_cfd_review, drl_cfd_review_part_2}.
To the best of our knowledge, no work has been done in applying DRL for CFD meshing, specifically.

While using DRL for CFD meshing is novel, applying ML techniques for faster solutions is not.
Works in operator learning\cite{https://doi.org/10.48550/arxiv.2205.13671}, surrogate modelling\cite{li_graph_2022, pant_deep_2021, https://doi.org/10.48550/arxiv.2005.06422}, and system identification \cite{meidani_data-driven_2021} offer alternative approaches to improve CFD simulations.
Operator learning and surrogate modelling offer different approaches to learn the temporal evolution of the system.
System identification is used to find new equations that describe the system at hand.
While these approaches show promise, meshing, specifically, is the focus of this work because it allows practitioners to use existing CFD solvers on the Navier-Stokes equations.


In this work, we develop a general purpose DRL framework to coarsen CFD meshes while maintaining accuracy in the calculation of coarse properties such as lift and drag.
Mesh coarsening is done by selecting and removing individual vertices in the mesh.
Proof of concept is demonstrated on 2D airfoils.
Our framework requires a single solve prior to training.
Interplay between different geometry, meshing, and simulation codes is often of practical concern.
Eliminating any dependence of our method on any particular code is of key importance for wide adoption.
This framework only requires modifying the mesh directly, which allows us to use standard CFD solvers to run subsequent simulations.
Lastly, our model makes no assumptions about the type of mesh, flow regime, solver, or dimensionality of the system.
This allows our method to be easily incorporated into CFD pipelines.

\section{Methods}

The MeshDQN framework is given below in figure \ref{fig:drl_workflow}.
After running an initial simulation, the pressures, velocities, coordinates, and edges are saved.
This data is then passed into the property calculation, where a coarse property, such as drag or lift, is calculated.
The reward is then calculated from the property.
\begin{figure}[ht]
    \centering
    \includegraphics[width=\linewidth]{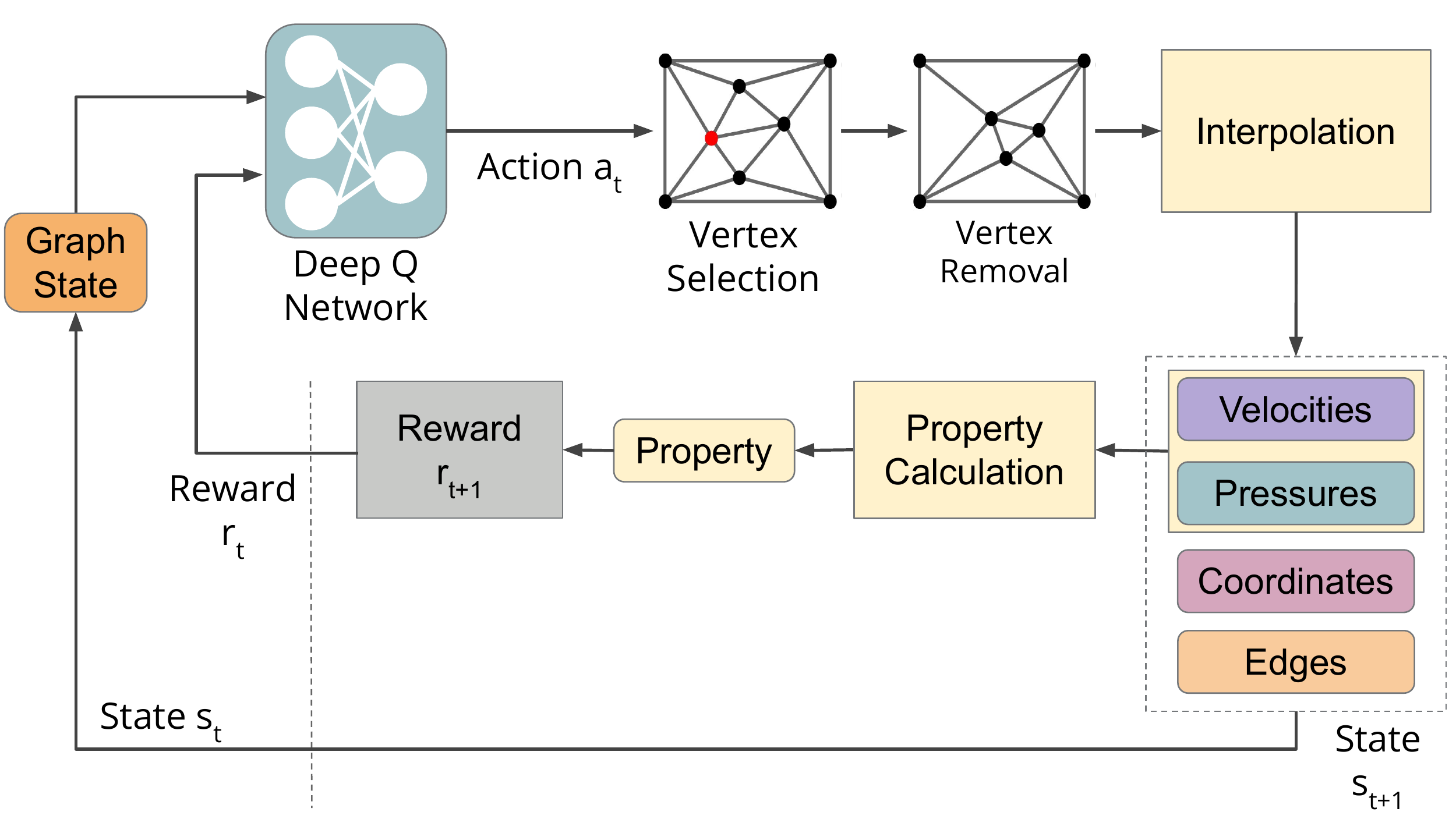}
    \caption{Each step begins by converting the current state $s_t$ into a graph representation as described above. The state and reward $r_t$ are passed into the DQN, a vertex is selected by action $a_t$ is chosen. The chosen vertex is then removed, the mesh is smoothed using local averaging roughly until convergence, where the velocity and pressure are interpolated to the new mesh, creating our next state $s_{t+1}$. The target property is calculated using the new velocities and pressures, where the reward $r_{t+1}$ is calculated. This workflow creates the tuple $(s_t, a_t, s_{t+1}, r_{t+1})$ necessary for Double DQN training.}
    \label{fig:drl_workflow}
\end{figure}

\subsection{Computational Fluid Dynamics}
\label{sec:cfd}
The goal of CFD is to solve the Navier-Stokes equations over a discrete mesh.
This work focuses on the incompressible NS equations with viscous flow:
\begin{equation}
    \rho\left(\frac{\partial\mathbf{u}}{\partial t} + u\cdot\nabla\mathbf{u}\right) = \nabla \cdot\sigma\left(\mathbf{u},p\right) + f, \quad\quad
    \nabla \cdot \mathbf{u} = 0
\end{equation}

Solving the NS equations numerically gives us pressure and velocity at each point.
The drag and lift are calculated from our solution velocity and pressure as:
\begin{equation}F_D = \int_{A}\left(\sigma\cdot \hat{n} \right)\cdot \hat{e}_x dS
\label{eq:drag}
\end{equation}
\begin{equation}F_L = \int_{A}\left(\sigma\cdot \hat{n} \right)\cdot \hat{e}_y dS
\label{eq:lift}
\end{equation}
Where $F_{D}$ and $F_{L}$ are drag force and lift force, respectively and $\sigma$ is the Cauchy stress tensor.

The meshes were generated using Delaunay Triangulation\cite{Barber96thequickhull} as part of GMSH\cite{gmsh}, with Meshio\cite{nico_schlomer_2018_1173116} used as an intermediary to convert mesh files into a usable format for the CFD solver. Simulations for this work were done over 5 seconds with the incremental pressure correction scheme (IPCS)\cite{goda_multistep_1979} implemented in Dolfin as part of FEniCS\cite{LoggEtal_10_2012, LoggWells2010}, based on the FEniCS Tutorial Volume 1\cite{langtangen_solving_nodate}.

\subsection{Deep Reinforcement Learning}
In reinforcement learning (RL), an agent interacts with an environment through actions.
The agent takes in a state $s$, selects an action $a$, which results in a new state $s'$.
Additionally, a reward $r$ is given after each action.

The goal in this case is to find an action selection policy such that reward is maximized.
One way of measuring the quality of a state-action pair, and learning this Q-function is known as Q learning.
The Q-function is defined in equation \ref{eq:q_learning}, where the reward is given based on taking action $a$ while in state $s$, and following the optimal policy afterwards.

\begin{equation}
    Q_{\pi}(s, a) \equiv \mathbb{E}\left[R_1 + \gamma R_2 + \cdots | S_0=s, A_0 = a, \pi\right]
    \label{eq:q_learning}
\end{equation}

Deep RL (DRL) specifically refers to the use of neural networks for policy estimation.
In this work, a graph neural network (GNN) is used to estimate the Q function, known as a Deep Q Netowrk (DQN).
Specifically, Double DQN is used, where one Q network performs action selection, and the other performs action evaluation, as in equation \ref{eq:dqn_select_eval}.
This is algorithm is used because it overestimates action values significantly less than DQN and is very easy to implement on top of DQN\cite{ddqn}.
\begin{equation}
    Y_{t}^{DoubleQ}\equiv R_{t+1} + \gamma Q_{eval}\left(S_{t+1}, \underset{a}{\text{argmax}}Q_{select}(S_{t+1}, a)\right)
    \label{eq:dqn_select_eval}
\end{equation}
The loss function, defined in equation \ref{eq:ddqn_loss}, is then a function of the expected action value, as estimated by $Q_{eval}$, and the actual selected action value, as measured by $Q_{select}$.
Gradients from the loss are propagated backwards to one network at a time, where the network being updated alternates every few episodes.

\begin{equation}
    \mathcal{L}^{DoubleQ} = \mathcal{L}\left(Y^{DoubleQ}_{t}, Q_{select}(s_t, a_t)\right)
    \label{eq:ddqn_loss}
\end{equation}

\subsection{Graph Neural Networks}

Meshes in CFD can be naturally interpreted as a graph $\mathcal{G} = (\mathcal{N}, \mathcal{E})$, where the nodes $\mathcal{N}$ are the mesh vertices, and the edges $\mathcal{E}$ are the connections between vertices.
By interpreting meshes as graphs, Graph Neural Networks (GNNs) become a natural model choice for learning in this context.
The structure of GNNs can be generally understood in three phases: aggregation, combination and readout.
The aggregation step aggregates information from nodes and their neighbors.
$$
\mathbf{a}_v^{(k)} = \text{AGGREGATE}^{(k)}\left(\left\{\mathbf{h}_u^{(k-1)}\colon u\in\mathcal{N}(v)\right\}\right)
$$
Once aggregated, the information is pooled, where popular approaches are mean and max pooling.
$$
\mathbf{h}_v^{(k)} = \text{POOL}^{(k)}\left(\mathbf{h}_v^{(k-1)}, \mathbf{a}_v^{(k)}\right)
$$
Finally, the readout operation is performed, often after several aggregation and combination steps.
Pooling operations are also popular for the readout step.
$$
\mathbf{h}_G = \text{READOUT}\left(\mathbf{h}_v^{(k)}\colon v \in \mathcal{G}\right)
$$
A more detailed description of GNNs can be found in a recent survey\cite{GNN}.

\begin{figure}[ht]
    \centering
    \includegraphics[width=\linewidth]{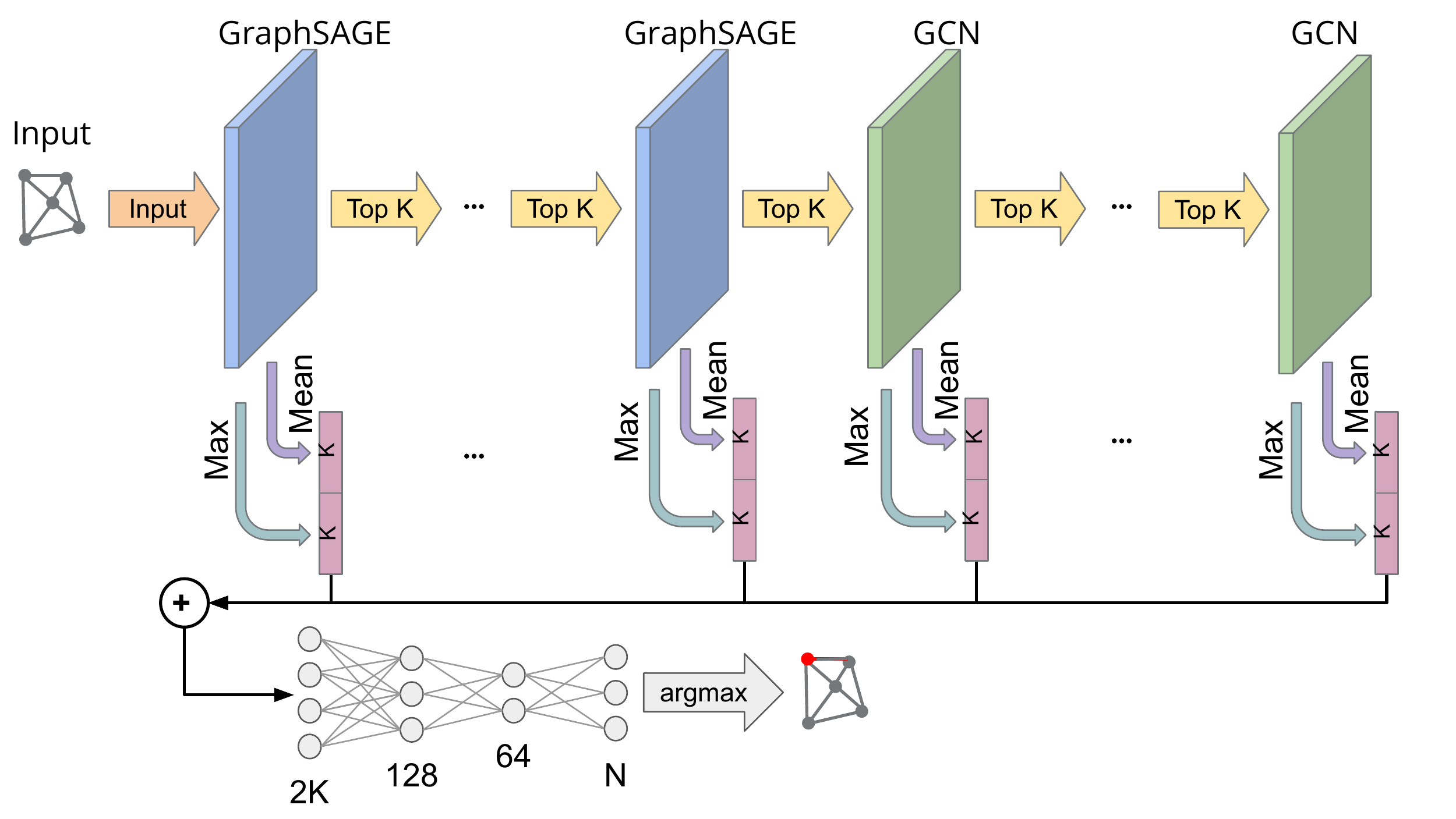}
    \caption{Deep Q Network used in MeshDQN. The graph representation of a mesh is passed into the first GraphSAGE layer. Top-K pooling is done after each graph layer. The output from all three GraphSAGE and GCN layers is then pooled with both mean pooling and max pooling into a concatenated vector of size 512 with skip connections. The output from each layer is then added together and passed into a fully-connected neural network classifier.}
    \label{fig:dqn_architecture}
\end{figure}

Node-level GNNs are used in this work because the values of interest, velocity and pressure, are defined at the nodes of the graph, and vertex selection is ultimately a node classification task.
The DQN architecture used in this work is based on AirfoilGCNN, seen in figure \ref{fig:dqn_architecture}, which is used for drag and lift prediction of airfoils in unstructured meshes\cite{Ogoke_2021}.
AirfoilGCNN makes use of GraphSAGE\cite{hamilton2017inductive} and Graph Convolutional\cite{GCN} layers as well as top-k pooling\cite{topk}.
AirfoilGCNN has been modified for use in MeshDQN with an additional GraphSAGE layer, an additional GCN layer.

GraphSAGE layers are specifically designed to generate embeddings for unseen nodes, and are very well suited in this case where actions can result in a new vertex being added to the state space.
Each GraphSAGE layer aggregates information from each vertex as well as its neighbors.
In this work, aggregation is done as:
\begin{equation}
    H^{(l)}_{\mathcal{N}(i)} = \text{MEAN}\left(H^{(l)}_j, \forall j\in \mathcal{N}(i)\right)
\end{equation}
And the next state is given by:
\begin{equation}
    H^{(l+1)}_i = \sigma\left(W_1^{(l)}H^{(l)}_i + W_2^{(l)}H^{(l)}_{\mathcal{N}(i)}\right)
\end{equation}
Here, $\mathcal{N}(i)$ is the set of all neighbors of vertex $i$, $\sigma$ is a nonlinearity, and $W^{(l)}_1$ and $W^{(l)}_2$ are trainable weights in each layer.

GCN layers have been shown to perform well in node classification tasks, and so are also suitable choice here.
GCN layers aggregate information in each layer, $l$, through the adjacency matrix with self-self connection $A$
\begin{equation}
    H^{(l+t)} = \sigma\left(D^{-1/2} A D^{-1/2}H^{(l)}W^{(l)}\right)
\end{equation}
where $D = \sum_j A_{ij}$, $W^{(l)}$ is a trainable weight matrix in each layer, and $\sigma$ is again a nonlinearity.

Top-K pooling is similar to max pooling, but selects the top-k fraction of data instead of the singular maximum value.

Skip connections are utilized in order to more effectively propagate information from each embedding into the final dense layers before classification.
Our model was developed using Pytorch Geometric\cite{Fey/Lenssen/2019}.

\subsection{DRL Workflow}
\begin{figure}[ht]
    \centering
    \includegraphics[width=\linewidth]{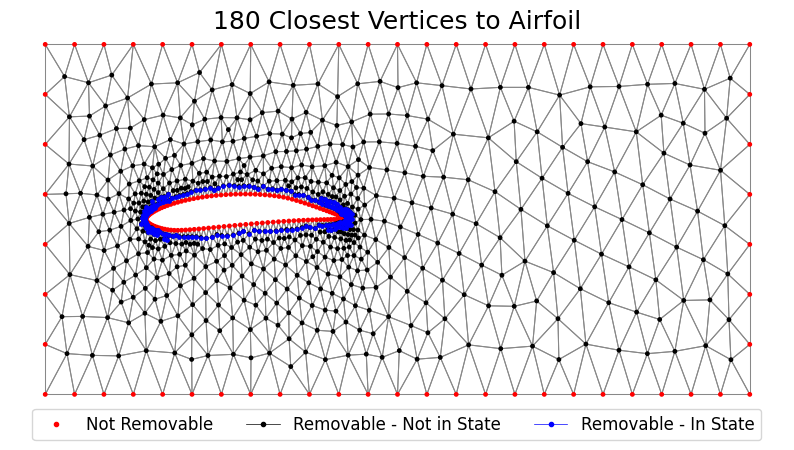}
    \caption{The initial state is given by the 180 vertices in the mesh closest to the airfoil, seen in blue. The remaining black vertices are added to the state as vertices are removed. Red vertices are the vertices on the boundary of the airfoil and simulation box. These are not removable and cannot be added to the state.}
    \label{fig:state}
\end{figure}

To begin our DRL training loop, we first run a simulation to calculate our target property using the IPCS scheme from section \ref{sec:cfd}.
Our target property is then computed using the equation for either drag (Eq. \ref{eq:drag}) or lift (Eq. \ref{eq:lift}) and is our ground truth value $p_{gt}$.
Intermediate and final solution states are stored as $\mathbf{u}_0$ and $\mathbf{p}_0$, respectively, which will be referred to as snapshots.
The number of snapshots to store can be tuned for the problem at hand.
This is used to construct our initial state, which is stored as a graph, $\mathcal{G} = \left(\mathcal{N}, \mathcal{E}\right)$.
Our state is defined as the $N$ closest vertices to the airfoil, seen in figure \ref{fig:state}, inspired by an approach for designing nanopores\cite{nanopore}.
Training was parallelized across 15 CPU cores using Ray, where 14 executed MeshDQN and the 15th acted as a server for DQN weights and updates.

\subsubsection{State and Action Space}
\label{sec:state_and_action_space}
The node-level information at each vertex is:
$$
V = \left[\mathbf{m}, \mathbf{u}, \mathbf{p}\right]
$$
Where $\mathbf{m}$ is the $x$ and $y$ coordinates of each mesh vertex, $\mathbf{u}$ is the $x$ and $y$ flow velocities at each vertex for each snapshot, $\mathbf{p}$ is the pressure at each vertex for each snapshot, and $\left[\cdot\right]$ indicates column-wise concatenation.
The edge index is simply the connectivity between vertices in our mesh and the edge attribute is the Euclidean distance between vertices.
The complete state space is given by $S$ which is a mixed continuous and discrete state space.
Our action space $A$ is all vertices in the current state, as well as an additional 'no removal' option that replaces the vertex nearest to the airfoil in the current state with the vertex nearest to the airfoil that is farther than all points in the current state.
The 'no removal' action stacks for a given episode, meaning that if the 'no removal' action is selected $N$ times, then for a state consisting of 200 vertices, the vertices are the  $N$th through $N + 200$th closest to the airfoil.

Before initialization and after each vertex removal, the mesh is smoothed with local averaging 50 times for approximate convergence.
Smoothing until convergence allows the full effect of each vertex removal to be realized immediately.
This helps decorrelate actions from each other, making the dynamics of the system simpler.

\subsubsection{Interpolation}
\label{sec:interpolation}
\begin{figure}[h]
    \centering
    \includegraphics[width=0.9\linewidth]{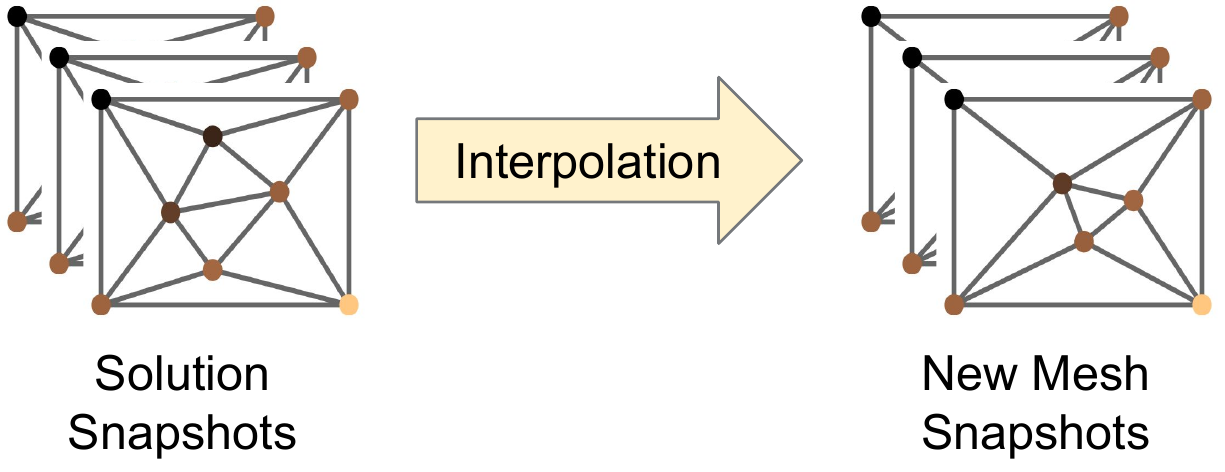}
    \caption{The initial pressures and velocities are interpolated onto the new mesh at each interpolation step.}
    \label{fig:interpolation}
\end{figure}

After vertex removal and smoothing is complete, the original snapshot values are interpolated onto the new mesh.
Interpolation is done using piecewise elements as supported by FEniCS.
Second order interpolation is done for velocities and first order is done for pressures.
This matches the element order of the finite element method used in these simulations, which has shown acceptable accuracy.
A toy example of interpolation is shown in figure \ref{fig:interpolation}.

\subsubsection{Reward Function}
\label{sec:reward_function}
Lastly, the reward function is defined in equation \ref{eq:reward} where a broken mesh or broken interpolation returns a negative reward and ends the episode.
The initial velocities and pressures are interpolated to the new mesh after each vertex removal and used for property estimation.
\begin{equation}
R = \begin{cases} 
      -1.0 & \text{Broken mesh,} \\ 
       & \text{broken interpolation} \\
      R_{property} + R_{time} & \text{Otherwise}
   \end{cases}
  \label{eq:reward}
\end{equation}
When the mesh and interpolation do not break, the reward is given by a property and time component, defined in equations \ref{eq:property_reward} and \ref{eq:time_reward}, respectively.
The factor $K$ is selected for an accuracy threshold of $0.1\%$, in line with thresholds found in literature\cite{ullah_cfd_2021}.
An error of $0.05\%$ yields a property reward of 0, and negative reward for worse accuracy.
When the accuracy threshold is exceeded after a vertex removal, the episode is ended.
Additionally, after a specified target number of vertices are removed the episode is ended.
This reward incentivizes the selection of vertices that reduce error the smallest amount, and begins to penalize vertices if their removal results in too large of a drop in accuracy.
The maximum property reward is 1, and the minimum is -1.
A more in-depth exploration of the property reward function is given in Appendix \ref{sec:reward_function_design}.
\begin{equation}
    R_{property} = 2\exp{\left(-K\left\lVert \frac{p_{gt} - p_{new}}{p_{gt}}\right\rVert_2\right)} - 1
    \label{eq:property_reward}
\end{equation}
The factor $K$ can then be computed analytically by setting $R_{property} = 0$ for an error of 0.0005 and solving for $K$: 
\begin{equation}
    K = \frac{-\log(0.5)}{0.0005} \approx 1386.29
\end{equation}
For the time reward, the factor $0.005$ is chosen so that the reward remains approximately constant after the removal of a vertex that results in a small, expected, drop in accuracy.
This incentivizes vertex removals, and without this, MeshDQN would be rewarded for selecting 'no removal' at every step.
The full workflow is given in figure \ref{fig:drl_workflow}.
\begin{equation}
    R_{time} = 0.005*\left(N_{gt} - N_{new}\right)
    \label{eq:time_reward}
\end{equation}



\section{Results}
\subsection{Mesh Improvement}
Drag, specifically is used as the target property in these experiments.
The YS930\cite{t_wing_1981} and AH93W145\cite{ah93w145} airfoils from the UIUC Airfoil Database\cite{selig_uiuc_1996} were used, with results in table \ref{tab:improvement_results}.
We use an accuracy threshold of 0.1\%\cite{ullah_cfd_2021} and aim for 5\% of vertices to be removed.
\begin{table}[h]
    \centering
    \begin{tabular}{c|ccc}
         Airfoil & Vertices Removed & Drag Error & Lift Error \\
         \hline
         YS930 & 5.023\%  & 0.039\% & 0.002\% \\
         AH93W145 &  5.019\% & 0.026\% & 1.782\% \\
    \end{tabular}
    \caption{MeshDQN is able to remove over 5\% of vertices for both airfoils, while inducing an error significantly smaller than our target threshold for our target property, drag. More lift error is induced in AH93W145, but this is well within the range of calculated lift values from standard meshes.}
    \label{tab:improvement_results}
\end{table}

\begin{figure}[ht]
    \centering
    \includegraphics[width=0.98\linewidth]{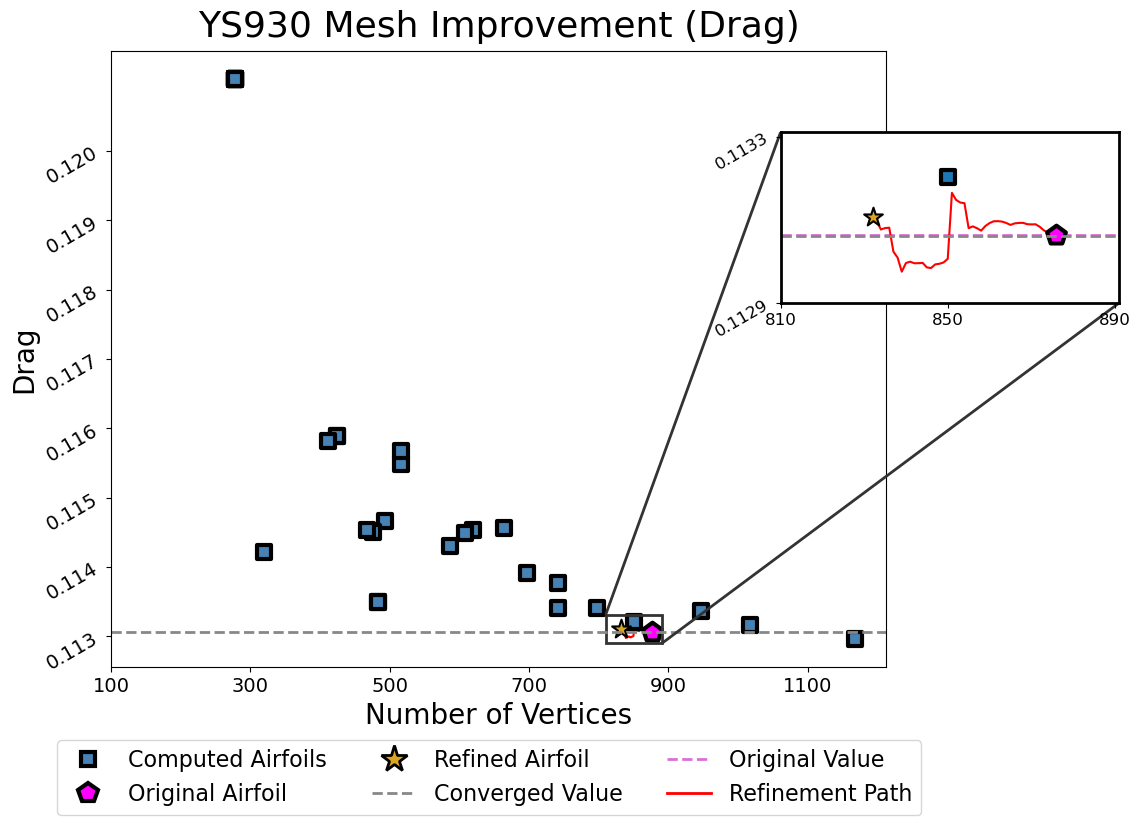}
    \caption{The trajectory of the computed drag is seen in red, where a full simulation is run after each vertex removal. We see drag is maintained after 5\% of vertices have been removed.}
    \label{fig:ys930_drag_trajectory}
\end{figure}

We see in figure \ref{fig:ys930_drag_trajectory} that MeshDQN is able to improve upon the selected airfoil, maintaining accuracy in the target property of drag for 44 vertices removed for the YS930 airfoil.
This accuracy is maintained past a mesh that was generated using Delaunay triangulation and smoothing.
Drag is also approximately maintained after each vertex removal, indicating MeshDQN is selecting vertices that maintain the drag at each step.
The initial and final YS930 airfoils are give in figure \ref{fig:ys930_before_and_after}.
Plots for the AH93W145 airfoil are given in Appendix \ref{sec:ah93w145_analysis}.
\begin{figure}[ht]
    \centering
    \includegraphics[width=0.98\linewidth]{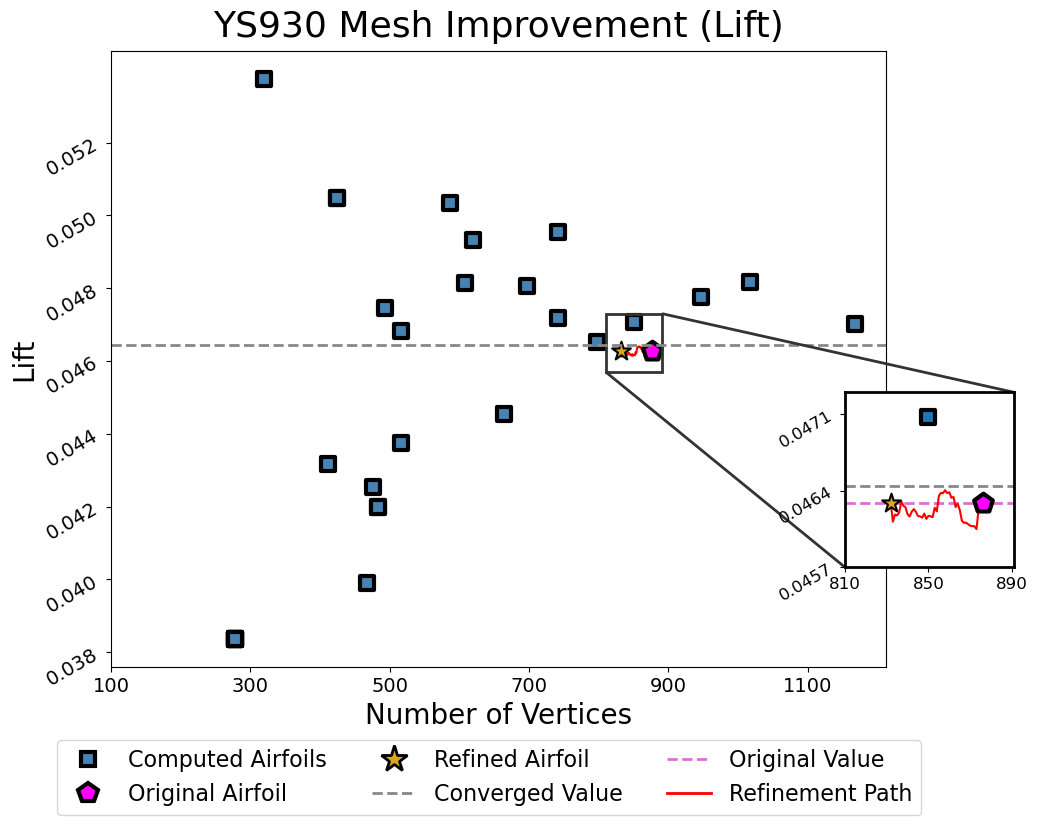}
    \caption{The trajectory of the computed lift is seen in red, where a full simulation is run after each vertex removal. We see lift is maintained fairly well after 5\% of vertices have been removed.}
    \label{fig:ys930_lift_trajectory}
\end{figure}

The initial and final meshes are given in figure \ref{fig:ys930_before_and_after}.
We see the mesh is sparser at the top and bottom surfaces of the airfoil, in the boundary flow layer where the flow is relatively simple.

\begin{figure}[ht]
    \raggedright
    a)
    \includegraphics[width=0.9\linewidth]{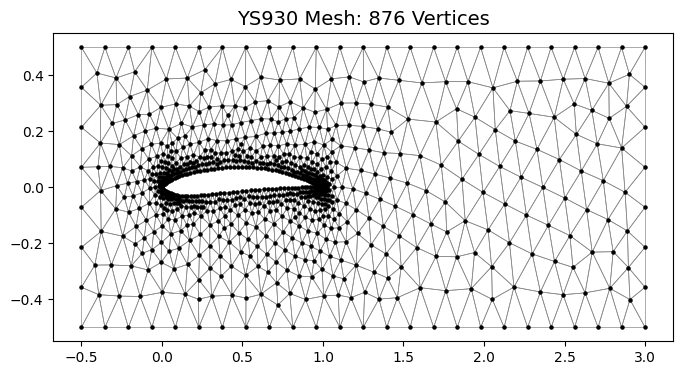}
    b)
    \includegraphics[width=0.9\linewidth]{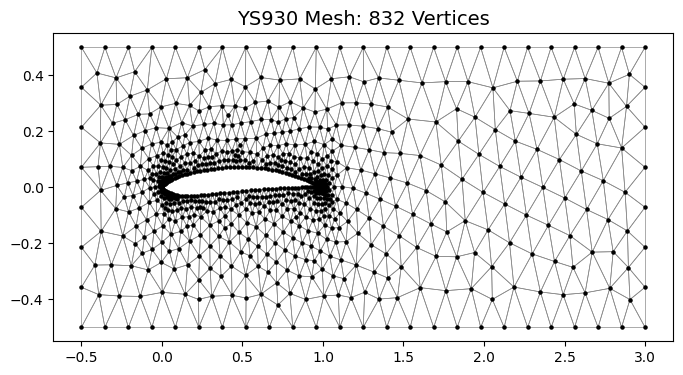}
    \caption{YS930 Airfoil with original, smoothed mesh on top (a), and final, improved mesh on the bottom (b) with unitless coordinates in the x and y directions. We see the agent has removed many vertices close to the airfoil, but has left vertices in complex flow regions at the leading and trailing edges of the airfoil.}
    \label{fig:ys930_before_and_after}
\end{figure}

\subsection{Interpolation Analysis}
\label{sec:interpolation_analysis}
\begin{figure*}[t]
    \centering
    \includegraphics[width=0.82\linewidth]{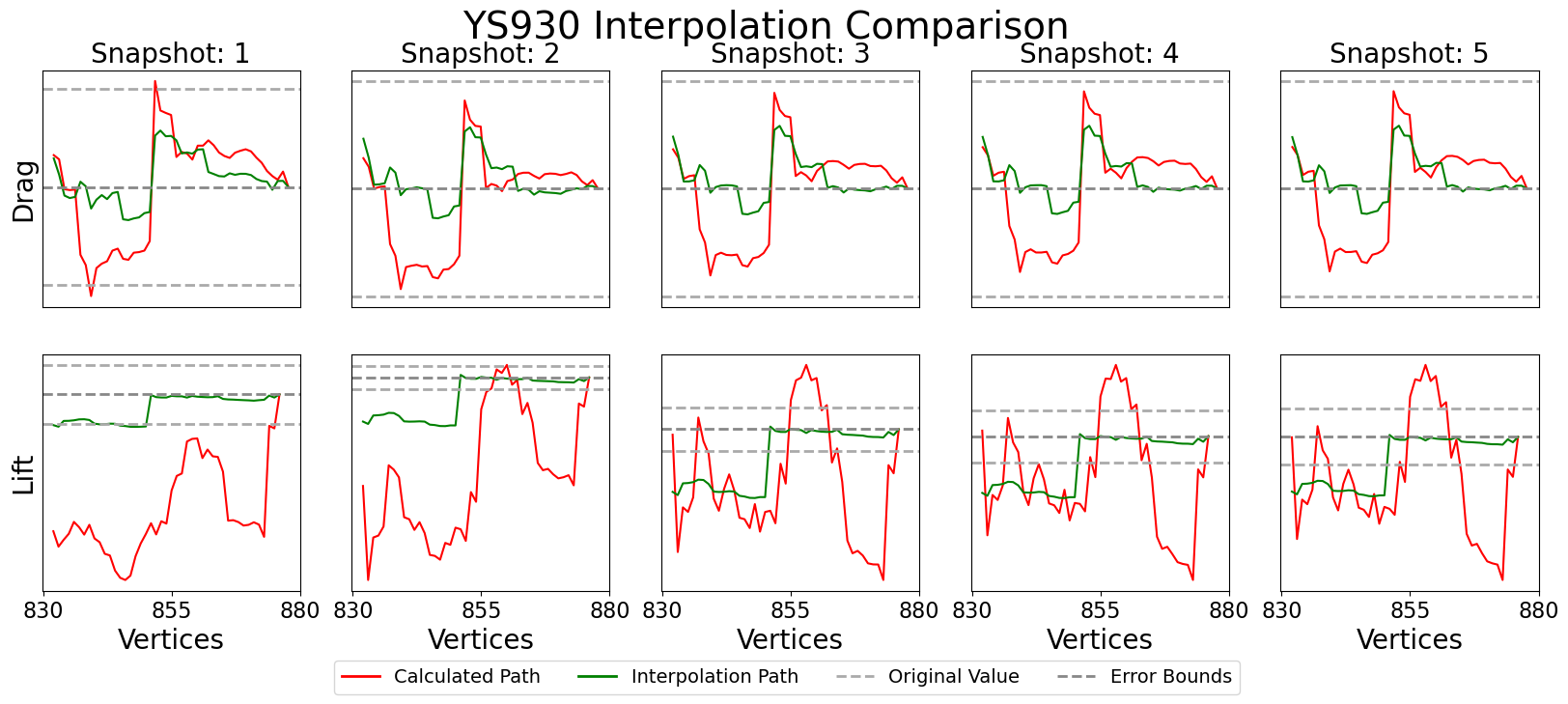}
    \caption{Comparing calculated (red) and interpolated (green) values for the drag and lift, we see good correlation for the target property of drag. Correlation between calculated and interpolated lift values is weaker.}
    \label{fig:ys930_comparison}
\end{figure*}
Interpolation is the key in allowing MeshDQN to quickly estimate target property values, and is of critical importance for accurate mesh improvement trajectories.
Interpolated and calculated values are compared in figure \ref{fig:ys930_comparison}.
The green trajectory is the interpolated drag and lift value after each vertex removal, and the red trajectory is the value calculated from a full numerical simulation.
In both figures, we see that the interpolated and calculated values correlate well with each other, generally showing shifts up and down in the same locations.
However, the magnitude of shift is not the same.
For lift, which was not used in training, there is little correlation between interpolated and calculated values.


\section{Conclusion}
We see MeshDQN has successfully taken fine meshes and selectively removed vertices while maintaining target property accuracy for 2D viscous, laminar flow for an unstructured mesh.
MeshDQN is able to improve the mesh for multiple airfoils, while making no assumptions about the solver scheme, mesh type, flow regime, or dimensionality.
While the current framework has demonstrated success at the given task, improved interpolation accuracy is of critical importance and is the current accuracy bottleneck.

While this work demonstrates a promising first step towards applying adaptive meshing to systems in CFD, there are many additional areas of CFD and meshing to explore.
A promising future direction of research would be to apply this framework to new mesh types, such as structured, block structured, and arbitrary polygon unstructured meshes.
Additionally, this work focused primarily on laminar flows, so applying this framework for turbulent flows remains an open problem.
Further, this framework could be adapted to 3D geometries.
Lastly, this framework was used simply for reducing the number of nodes in a fine mesh.
In order to fully adapt a mesh to a particular problem, adding vertices to the mesh is necessary for refinement in regions of complex flow with high Reynolds number.
This would also allow practitioners to begin with a relatively coarse mesh easy to generate mesh and adaptively refine the mesh to a given problem, making the CFD development significantly easier.

\section*{Author Contributions}
Cooper Lorsung: conceptualization (equal); data curation, formal analysis; methodology (lead); software; validation (supporting); visualization; writing - original draft preparation; writing - review and editing (lead)

Amir Barati Farimani: conceptualization (equal); resources; methodology (supporting); resources; validation (supporting); supervision; writing - review and editing (supporting)

\section*{Acknowledgements}
CL acknowledges Francis Ogoke for providing starter simulation code and helping with debugging.
CL acknowledges Zhonglin Cao for the idea of using $N$ closest vertices to the airfoil.

This material is based upon work supported by the National Science Fountation under Grant No. 1953222

\section*{Author Declaration}
The authors have no conflicts to disclose.

\section*{Data Availability}
MeshDQN code and data are available at: \href{https://github.com/BaratiLab/MeshDQN}{https://github.com/BaratiLab/MeshDQN}

\clearpage
\newpage
\appendix
\section{Reward Function Design}
\label{sec:reward_function_design}

Reward function design is a critical component of the MeshDQN framework.
Selecting a good $K$ value, as in equation \ref{eq:property_reward}, determines how strongly a vertex removal with high error, and in turn, allows MeshDQN to distinguish between low and high error vertex removals.
Selecting too small of a $K$ value leads to rewards that are too close together to reliably learn which vertex removals lead to low and high error.
On the other hand, selecting too large of a $K$ value causes issues in the other direction, where too many vertex removals are penalized, even with fairly small error.
Selecting a proper $K$ value is therefore a balancing act.

\begin{figure}[h]
    \centering
    \includegraphics[width=\linewidth]{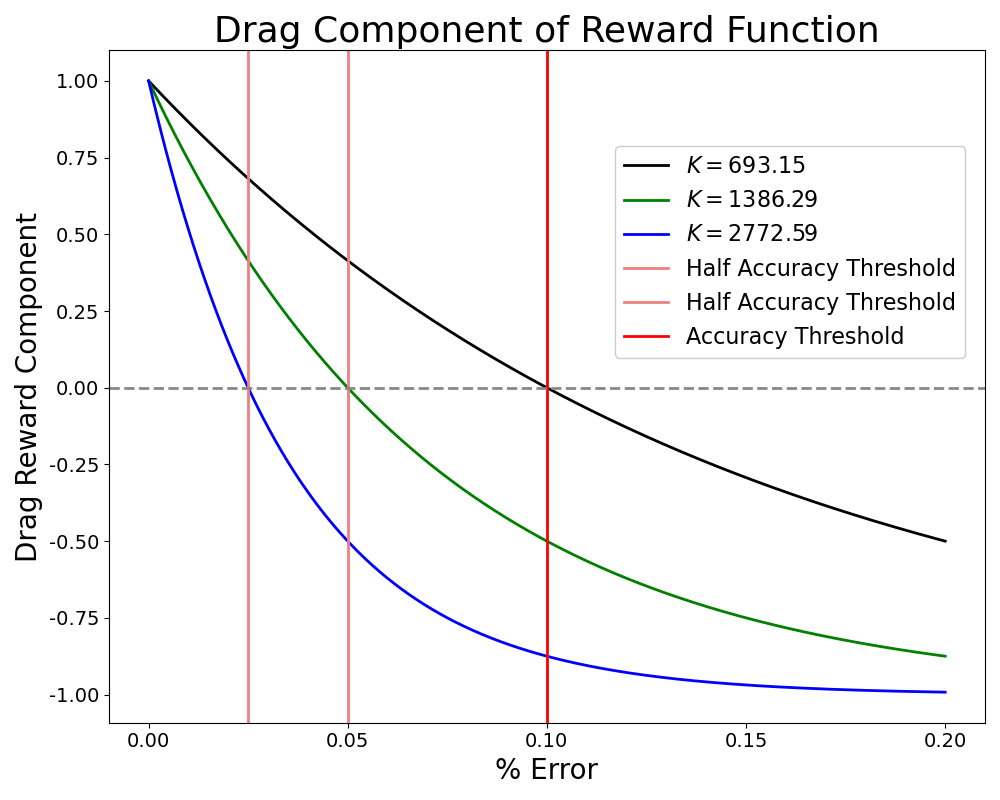}
    \caption{
        Different choices for property reward $K$ have zero reward at different locations relative to the accuracy threshold.
        Intuitively, penalizing vertex removals before we get to the error threshold disincentivizes vertex removals that accumulate too much error.
    }
    \label{fig:drag_reward}
\end{figure}

As mentioned in section \ref{sec:reward_function}, intuition guides us to select $K$ based on where we have $0$ reward.
This allows us to set an penalty threshold, above which we penalize vertex removals, and below which we incentivize them.
We see in figure \ref{fig:drag_reward} three different options.
We can set 0 reward at the total error threshold, halfway to the error threshold, or a quarter of the way to the error threshold.
Empirically, we have found setting the penalty threshold halfway to the way to the error threshold works well, leading to a variety of actions selected.
Setting the penalty threshold higher does not generally provide a strong enough penalty for MeshDQN to reliably distinguish between actions, often times leading to mode collapse and a single action being selected repeatedly.
Setting the threshold lower often leads to all removals being penalized too strongly, and MeshDQN learns to only select 'no removal'.

%

\section{AH93W145 Analysis}
\label{sec:ah93w145_analysis}
\begin{figure}[ht]
    \centering
    \includegraphics[width=0.98\linewidth]{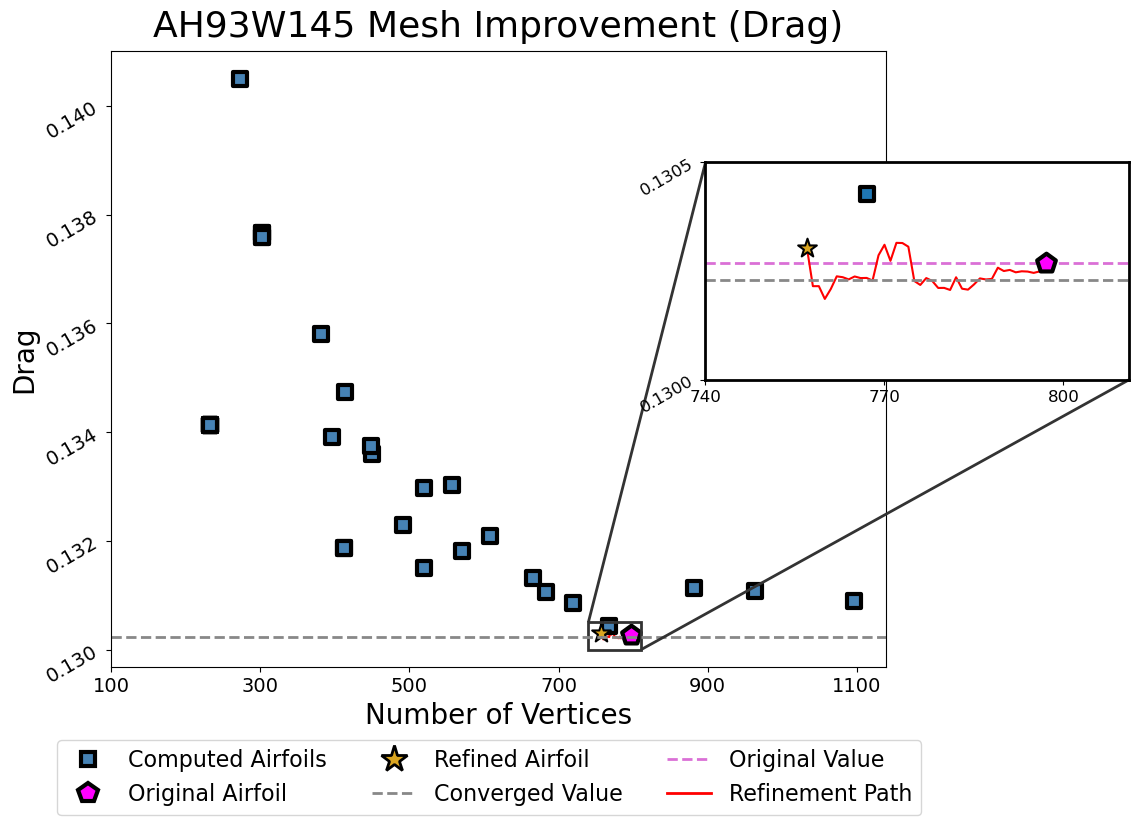}
    \caption{The trajectory of the computed drag and lift are seen in red, where a full simulation is run after each vertex removal. We see drag is maintained after 5\% of vertices have been removed. The calculated lift value is also preserved despite not being used in training. The converged value is the calculated drag or lift at the finest mesh used.}
    \label{fig:ah93w145_drag_trajectory}
\end{figure}

\begin{figure}[ht]
    \centering
    \includegraphics[width=0.98\linewidth]{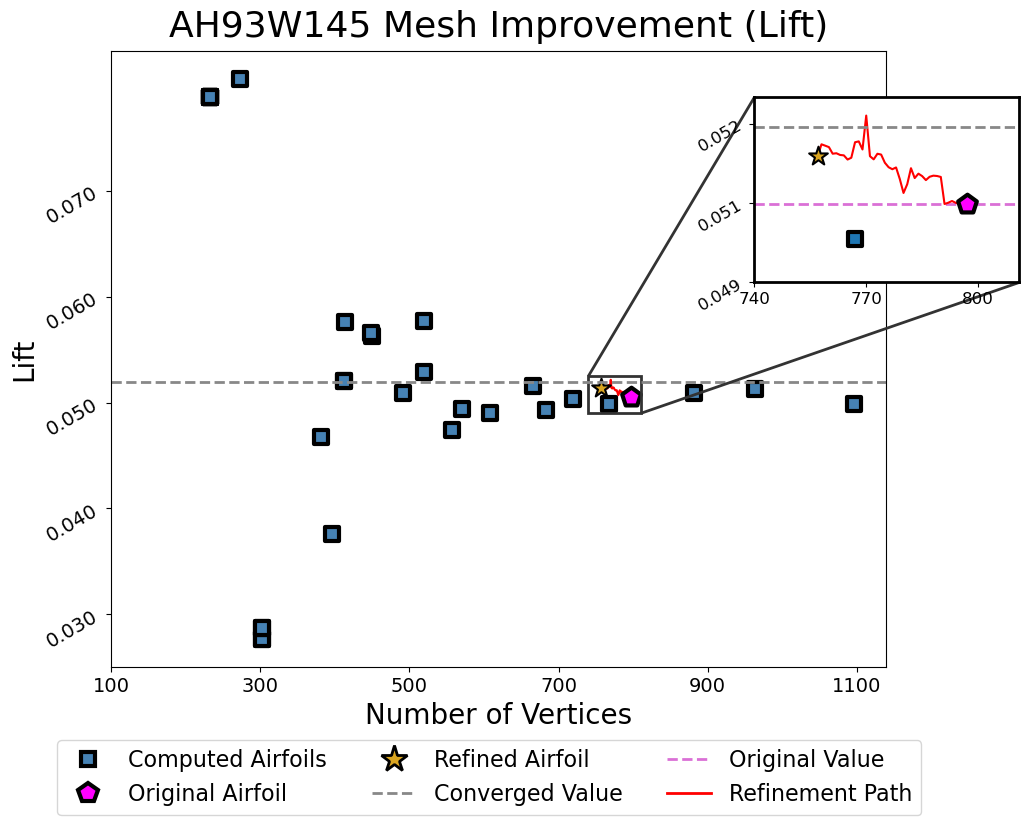}
    \caption{The trajectory of the computed drag and lift are seen in red, where a full simulation is run after each vertex removal. We see drag is maintained after 5\% of vertices have been removed. The calculated lift value is also preserved despite not being used in training. The converged value is the calculated drag or lift at the finest mesh used.}
    \label{fig:ah93w145_lift_trajectory}
\end{figure}

\begin{figure}[ht]
    \raggedright
    a)
    \includegraphics[width=\linewidth]{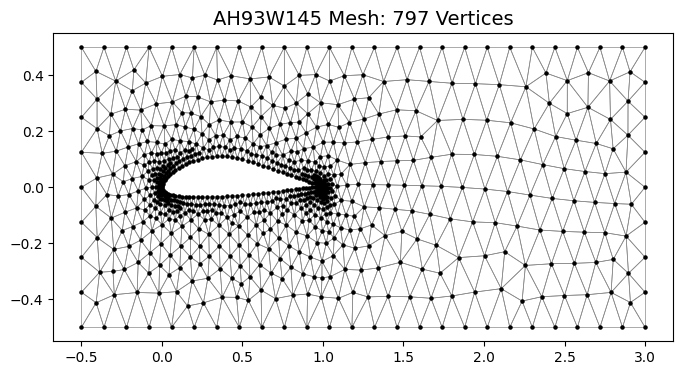}
    b)
    \includegraphics[width=\linewidth]{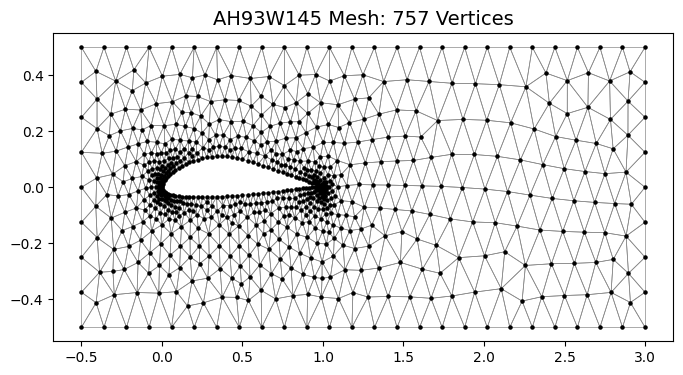}
    \caption{AH93W145 airfoil with original, smoothed mesh on top (a), and final, improved mesh on the bottom (b) with unitless coordinates in the x and y directions. We see the agent has removed many vertices close to the airfoil, but has left vertices in complex flow regions at the leading and trailing edges of the airfoil.}
    \label{fig:ah93w145_before_and_after}
\end{figure}

Further analysis of the AH93W145 airfoil is given here.
We have the drag improvement in figure \ref{fig:ah93w145_drag_trajectory}, where we see the drag is improved past a generated meshing.
Lift improvement is given in figure \ref{fig:ah93w145_lift_trajectory}, where we see the calculated lift value is not maintained as well, but still remains well within the region of other generated meshes.
The initial and final meshes are given in figure \ref{fig:ah93w145_before_and_after}, where we see the mesh at the top and bottom faces of the airfoil have been slightly coarsened.
\begin{figure}[t]
    \centering
    \includegraphics[width=0.8\linewidth]{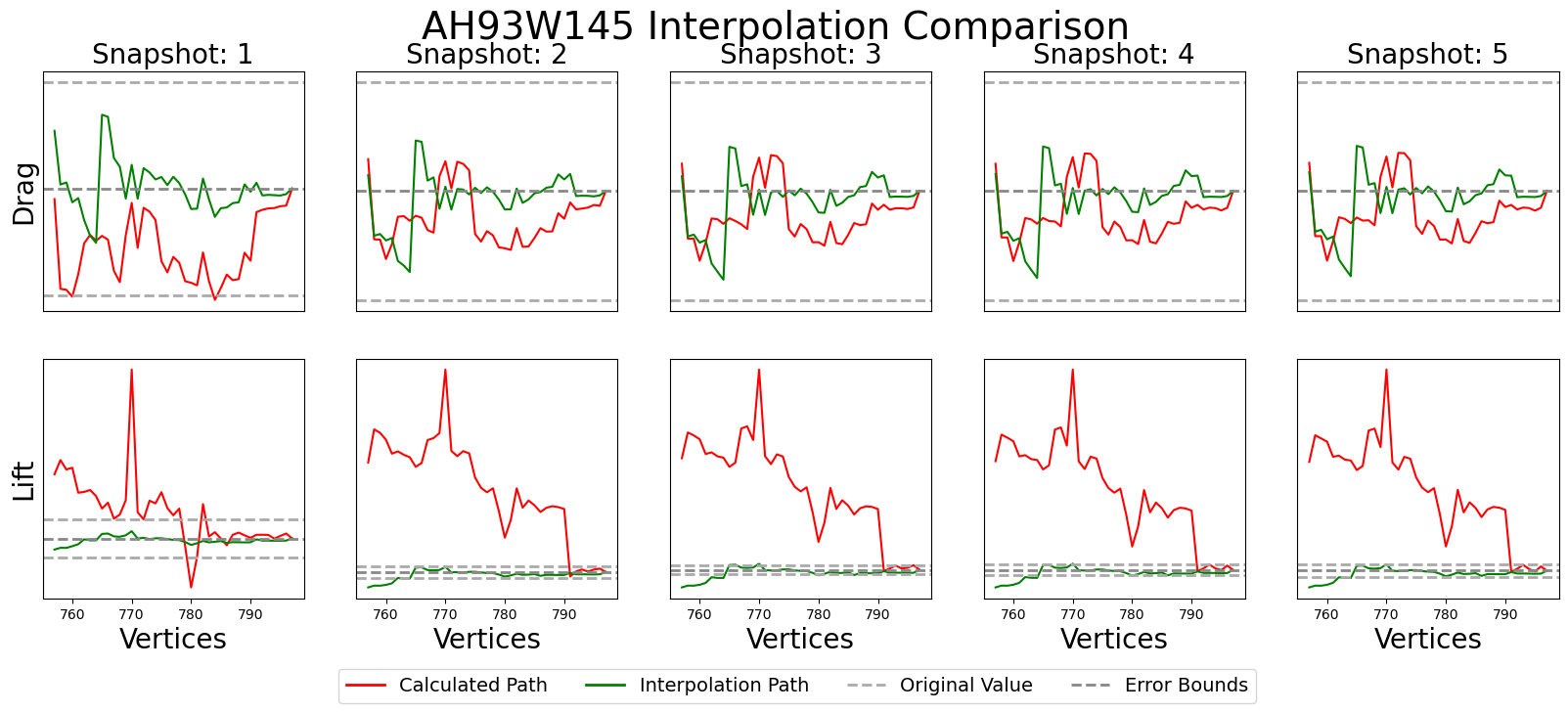}
    \caption{Comparing calculated (red) and interpolated (green) values for the drag and lift, we see good correlation for the target property of drag. Correlation between calculated and interpolated lift values is weaker, especially for later snapshots.}
    \label{fig:ah93w145_comparison}
\end{figure}


\section{Simulation and Training Configuration}
\label{sec:sim_and_train}
\begin{figure}[h]
    \centering
    \includegraphics[width=\linewidth]{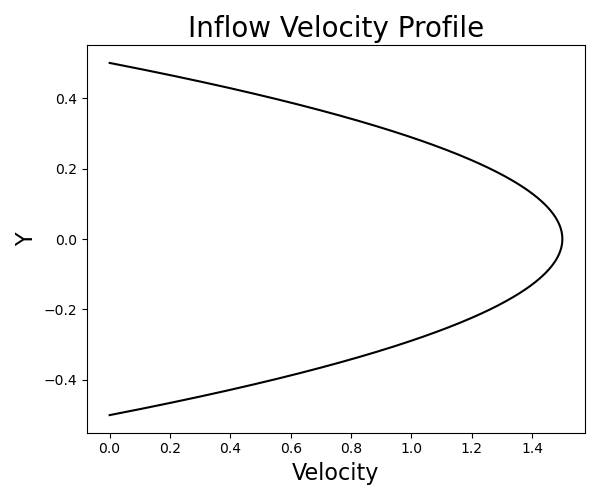}
    \caption{The inflow velocity is based on the well known Poiseuille flow.}
    \label{fig:inflow_profile}
\end{figure}

The flow solver was configured with no-slip boundary conditions on the airfoil, top and bottom walls.
Zero pressure was used at the outlet.
The inflow velocity profile is constant with respect to time and given in figure \ref{fig:inflow_profile}.
The simulation was done with unitless equations.
Viscosity was set to 0.001, with density of 1, which gives us a Reynolds number of 1800.
The timestep was set to 0.001, for 5000 timesteps, for a total of a five second simulation.
The discount factor $\gamma$ was set to 1.
The Adam optimizer\cite{DBLP:journals/corr/KingmaB14} was used with a learning rate of 0.0005 and no weight decay.
The weights and biases of each layer were initialized with Xavier normal initializer\cite{pmlr-v9-glorot10a} using a gain of 0.9.
A convolutional width of 128 was used for each GraphSAGE and GCN layer.
Two GraphSAGE and GCN layers were used.
Additionally, training was parallelized using Ray.
Five evenly spaces snapshots were used at timestep 1000, 2000, 3000, 4000, and 5000.

\clearpage
\newpage
\section*{References}
\bibliographystyle{imr}
\setstretch{2}
\bibliography{the}

\end{document}